\documentclass[10pt]{article}

\usepackage{geometry}
\usepackage{microtype}
\usepackage{amsmath, amssymb, amsthm, mathtools}
\usepackage{graphicx}
\usepackage{subfigure}
\usepackage{booktabs}
\usepackage{multirow}
\usepackage{algorithm}
\usepackage{algorithmic}
\usepackage{float}
\usepackage{tikz}
\usetikzlibrary{calc, arrows.meta, positioning}
\usepackage{hyperref}
\hypersetup{
    colorlinks=true,
    linkcolor=blue,
    citecolor=blue,
    urlcolor=blue,
}
\usepackage[capitalize,noabbrev]{cleveref}  
\usepackage[numbers]{natbib}               

\DeclareUnicodeCharacter{2009}{\,}      
\DeclareUnicodeCharacter{2264}{$\le$}
\DeclareUnicodeCharacter{2265}{$\ge$}
\DeclareUnicodeCharacter{2248}{$\approx$}
\DeclareUnicodeCharacter{00D7}{$\times$}
\DeclareUnicodeCharacter{21D2}{$\Rightarrow$}
\DeclareUnicodeCharacter{2191}{$\uparrow$}
\DeclareUnicodeCharacter{2193}{$\downarrow$}

\theoremstyle{plain}
\newtheorem{theorem}{Theorem}[section]

\newtheorem{proposition}[theorem]{Proposition}
\theoremstyle{definition}
\newtheorem{definition}{Definition}[section]

\title{\textbf{RN-F: A Novel Approach for Mitigating Contaminated Data in Large Language Models}}

\author{
    Le Vu Anh$^{1,2,3}$ \quad
    Dinh Duc Nha Nguyen$^{1,3}$ \quad
    Phi Long Nguyen$^{2}$ \\[0.5em]
    $^1$TinyAI Research \\
    $^2$Center for Environmental Intelligence, VinUniversity, Hanoi, Vietnam \\
    $^3$College of Engineering \& Computer Science, VinUniversity, Hanoi, Vietnam \\[0.5em]
    \texttt{nha.ndd@vinuni.edu.vn}
}

\date{\today}

\begin{document}
\maketitle

\begin{center}
\end{center}
\vspace{1em}

\begin{abstract}
Large Language Models (LLMs) have become foundational in modern artificial intelligence, powering a wide range of applications from code generation and virtual assistants to scientific research and enterprise automation. However, concerns about data contamination—where test data overlaps with training data—have raised serious questions about the reliability of these applications. Despite awareness of this issue, existing methods fall short in effectively identifying or mitigating contamination. In this paper, we propose Residual-Noise Fingerprinting (RN-F), a novel framework for detecting contaminated data in LLMs. RN-F is a single-pass, gradient-free detection method that leverages residual signal patterns without introducing additional floating-point operations. Our approach is lightweight, model-agnostic, and efficient. We evaluate RN-F on multiple LLMs across various contaminated datasets and show that it consistently outperforms existing state-of-the-art methods, achieving performance improvements of up to \textbf{10.5\%} in contamination detection metrics.
\end{abstract}

\section{Introduction}\label{sec:intro}

Large Language Models (LLMs) have become foundational tools in modern artificial intelligence, supporting diverse applications such as code generation, scientific discovery, enterprise automation, and intelligent assistants~\cite{tang2024code, boiko2023autonomous, aggarwal2025scriptsmith, qin2024toolllm}. Their remarkable performance on various benchmarks~\cite{du2024evaluating, liu2024agentbench} has driven rapid adoption across both academia and industry. However, this progress has raised serious concerns about the validity of benchmark results due to potential data contamination~\cite{deng2024investigating}. As LLMs are often trained on massive, uncurated internet-scale datasets, it becomes increasingly likely that test data may overlap with training data—either directly or through paraphrased or synthetic forms—thereby artificially inflating model performance and misleading evaluations.

Data contamination refers to the inadvertent inclusion of benchmark or evaluation data within a model’s training corpus~\cite{dong2024generalization, deng2024investigating}. This phenomenon results in LLMs memorizing answers rather than generalizing from learned patterns, thereby jeopardizing the trustworthiness of performance evaluations. Contaminated models can exhibit strong results on known benchmarks while failing to perform reliably on unseen or real-world tasks~\cite{li2024task}. This issue threatens both scientific progress and application safety, highlighting an urgent need for effective, scalable, and model-agnostic methods to detect and mitigate contamination in LLMs.

Existing state-of-the-art approaches~\cite{dong2024generalization, shen2025bait, dekoninck2024constat} to contamination detection typically rely on access to internal model parameters, output probabilities, or multiple reference datasets. These methods often require repeated model evaluations or comparisons across rephrased benchmarks, synthetic data, or assumed-clean samples. While effective in certain controlled settings, they are generally resource-intensive, difficult to scale, and impractical for quantized or edge-deployed models. Furthermore, many current methods struggle to detect subtle or implicit forms of contamination, especially when training data is opaque or continuously evolving.

In this paper, we propose Residual-Noise Fingerprinting (RN-F), a novel, efficient framework for detecting data contamination in LLMs. RN-F exploits a simple yet powerful signal—the quantization residual—defined as the per-layer difference between full-precision and quantized model activations. By observing how this residual behaves differently on contaminated inputs versus clean data, RN-F serves as a lightweight, gradient-free, and single-pass anomaly detector. It requires no access to internal gradients or model retraining and is suitable for deployment on low-resource devices. Extensive experiments across multiple models and datasets show that RN-F outperforms existing methods, achieving up to 10.5\% higher detection performance while maintaining minimal computational overhead.

This work offers three primary contributions:
\begin{enumerate}
    \item We introduce Residual-Noise Fingerprinting (RN-F), a pioneering approach that leverages the quantization residual—the per-layer difference between full-precision and int4 activations—as an anomaly signal. To the best of our knowledge, RN-F is the first framework to repurpose quantization artefacts as a tool for contamination detection in LLMs and other compact models, especially under severe resource constraints.
    \item We provide a rigorous statistical characterization of quantization residuals, proving that clean inputs exhibit sub-Gaussian tails while contaminated inputs induce a bounded mean shift. This theoretical analysis enables provable guarantees for false positive and false negative rates, even with a limited calibration buffer.
    \item We evaluate RN-F on compact models across tabular, language, and image tasks, demonstrating superior performance over state-of-the-art methods while maintaining a lightweight, single-pass, gradient-free setup with minimal calibration.
\end{enumerate}

The rest of the paper unfolds as follows.  Section~\ref{sec:related} positions our work among existing defences. Section~\ref{sec:method} formalises the quantization residual, and Section~\ref{sec:maindetector} turns it into the RN-F algorithm. Results and ablations, followed by a discussion of limitations, appear in Section~\ref{sec:setup}. Technical proofs and additional plots live in the Appendix.

\section{Related Work}\label{sec:related}

Recent advances in large language models (LLMs) have prompted numerous studies on detecting data contamination and backdoor vulnerabilities. These works explore various detection strategies, from statistical analysis to probing internal representations. Below, we summarize key contributions in this space.

\par
This work~\cite{golchin2024time} proposes a guided method to detect contamination by prompting LLMs with dataset-specific cues, comparing outputs via ROUGE-L and BLEURT. It scales from instance- to partition-level but assumes near-exact text matches, missing semantic paraphrases. The authors~\cite{yan2024rethinking} study poisoning intensity effects and find detectors fail under both strong and weak poisoning, revealing brittleness under varying attack strengths. This survey~\cite{fu2025does} reviews 50 detection methods by their assumptions, showing many fail under distribution shifts where memorization assumptions break down. The authors~\cite{samuel2025towards} benchmark five detectors on four LLMs and eight datasets, noting performance drops with instruction-tuned models and complex prompts. RECALL~\cite{xie2024recall} uses changes in log-likelihoods under prefix perturbation for membership inference, achieving strong results but requiring token-level access, limiting black-box use. This method~\cite{liu2024probing} uses hidden layer activations with a probe classifier to detect training data, performing well but needing proxy models and internal access. DC-PDD~\cite{zhang2024dc} calibrates token probabilities using divergence from expected frequencies, reducing false positives but depending on raw token access and vocabulary stability. ConStat~\cite{dekoninck2024constat} defines contamination as performance inflation, using statistical comparisons and p-values but needs curated benchmarks and assumes task generalization. BAIT~\cite{shen2025bait} inverts target sequences to detect backdoors via generation probabilities; effective for generative tasks but reliant on specific causal structures. CDD~\cite{dong2024generalization} identifies peaked output distributions as signs of memorization, performing well and efficiently, though its confidence-based signal may not generalize.

\par
\textbf{Our Observation.} Despite promising results, most existing methods depend on strong assumptions: access to logits, full-precision models, or curated references. Many break under distribution shifts, quantization, or black-box constraints. This highlights the need for lightweight, generalizable methods like RN-F that require minimal assumptions and operate efficiently across settings.

\section{Preliminaries}\label{sec:method}

\subsection{Problem set-up}
Let \(f_{\theta}\colon \mathbb{R}^{m}\!\to\!\mathbb{R}^{C}\) denote a \emph{frozen} fp16 network with \(L\) layers, and let
\(\hat f^{(4)}_{\theta}\) be the same network after post-training 4-bit quantization (PTQ) using a uniform step
\(2q\).  We write \(h_\ell(x)\in\mathbb{R}^{d_\ell}\) for the fp16 activation of layer~\(\ell\) and
\(\hat h_\ell(x)\) for its int4 counterpart.

\begin{definition}[Layer-wise residual]\label{def:res}
For an input \(x\) and a layer of width \(d_\ell\), the
\emph{quantization residual}
is the mean absolute deviation between fp16 and int4 activations:
\[
r_\ell(x)\;=\;\frac1{d_\ell}\,
              \Bigl\|h_\ell(x)-\hat h_\ell(x)\Bigr\|_1.
\]
\end{definition}

Intuitively, \(r_\ell(x)\) measures how far the int4 grid has to “snap’’ the
fp16 activation vector.  We aggregate over layers by
\(r_{\max}(x)=\max_{\ell} r_\ell(x)\); other norms (e.g.\ sum or average)
behave similarly and are analysed in Appendix A.1.

\subsection{Why does the residual separate clean from tainted inputs?}

\paragraph{Uniform quantization as structured noise.}
PTQ rounds each coordinate \(h_{\ell,i}\) to the nearest grid point
\(g\in q\mathbb{Z}\).  The rounding error
\(\varepsilon_{i}= \hat h_{\ell,i}-h_{\ell,i}\)
is thus uniformly distributed in \([{-}q,q]\) \emph{conditioned} on the event
that \(h_{\ell,i}\) lands in a high-density cell.  
For in-distribution (ID) data, successive activations visit those high-density
cells almost uniformly, so the errors
\(\{\varepsilon_i\}\) have mean~0 and cancel out:
\(\mathbb{E}[r_\ell(x)]\approx0.\)

\begin{figure}[t]
  \centering
  \includegraphics[width=\columnwidth]{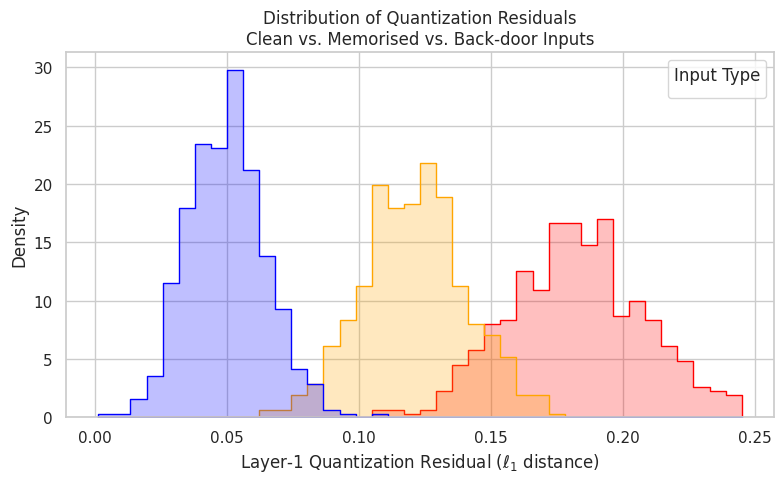}
  \vspace{-0.8em}
  \caption{Layer-1 quantization residuals reveal anomalous inputs.  
           Clean data cluster near zero (blue), whereas memorised (orange) and back-door (red) samples shift the distribution to the right.}
  \label{fig:residual_hist}
  \vspace{-1em}
\end{figure}

\paragraph{Contaminated inputs break the symmetry.}
A memorised or back-doored input drives the fp16 activations to
rarely-visited regions of feature space.  
In those sparse cells the quantizer no longer sees symmetric neighbours; the
rounding error is biased in one direction, so the absolute residual
\(r_\ell(x)\) spikes.  
Empirically (Figure \ref{fig:residual_hist}) even a single layer suffices to
separate ID and anomalous inputs, but in RN-F we keep all layers for stronger
statistical power.

\subsection{Distributional analysis}\label{sec:theory}

We next formalise the intuition under a mild smoothness assumption.

\begin{proposition}[Sub-Gaussian tail on ID data]\label{prop:sub}
Assume each coordinate rounding error
\(\varepsilon\sim\mathrm{Unif}[-q,q]\)
and that the layer mapping \(x\mapsto h_\ell(x)\) is \(K\)-Lipschitz.
Then for any ID input \(x\) and any \(\tau>0\),
\[
    \Pr\Bigl(|r_\ell(x)-\mu_\ell|\;>\;\tau\Bigr)
    \;\le\;2\exp\!\Bigl[-\,d_\ell\tau^{2}\big/(2q^{2}K^{2})\Bigr],
\]
where \(\mu_\ell=\mathbb{E}[r_\ell]\).
\end{proposition}

The bound follows from Hoeffding on the sum of
\(|\varepsilon_i|\), each of which is sub-Gaussian with parameter
\(q^2/3\).  A wider layer (\(d_\ell\!\uparrow\)) therefore yields a sharper
concentration, a property we exploit when setting thresholds.

\begin{theorem}[Instance-level detection guarantee]\label{thm:det}
Let contaminated inputs shift the mean residual by at least \(\Delta>0\),
i.e.\ \(\mathbb{E}[r_\ell(x)\mid x\!\in\!\text{tainted}]-\mu_\ell\ge\Delta\)
for some layer \(\ell\).  
Calibrate RN-F with \(n\) clean samples and choose the
threshold \(\tau\) as the
\(1-\alpha\) empirical quantile of \(r_\ell\).
If
\[
    n\;\ge\;8\,q^{2}K^{2}\,\frac{\log(2/\epsilon)}{\Delta^{2}},
\]
then RN-F achieves \(\text{FPR}\le\epsilon\) and \(\text{FNR}\le\epsilon\).
\end{theorem}

\paragraph{Proof sketch.}
With \(n\) IID calibration points the empirical mean
\(\hat\mu_\ell\) deviates from \(\mu_\ell\) by at most
\(\Delta/2\) with probability \(1-\epsilon\) (Hoeffding).  
Setting \(\tau=\hat\mu_\ell+\Delta/2\) then ensures
clean inputs lie below the threshold (FPR) and contaminated inputs lie above
it (FNR) with the same confidence.  Detailed steps are given in Appendix A.

\paragraph{Corollary (layer max).}
Because \(r_{\max}(x)=\max_\ell r_\ell(x)\) is the point-wise maximum of sub-Gaussian variables, it inherits a sub-Weibull tail with parameter
\(2^{-1}\).  
Consequently, bounding \(r_{\max}\) delivers a union-free test across layers without an additional Bonferroni penalty.

\subsection{Practical calibration}
RN-F uses \(n=512\) clean points—well within the memory of a Colab T4—to estimate \(\hat\mu_\ell\) for every layer. The threshold \(\tau\) is then fixed once and reused across
all future inferences; Section~\ref{sec:experiments} shows that AUC saturates long before \(n=512\), confirming the finite-sample guarantee in Theorem~\ref{thm:det}.

\paragraph{Connection to Fisher information.}
For small \(q\) the quantization error acts as an isotropic perturbation, so
\(\Sigma_\ell=\operatorname{Cov}[\hat h_\ell-h_\ell]\approx q^{2}I\).  
Hence the Mahalanobis distance \(r_\ell(x)\) is, up to scaling, the outer
product \(g_\ell(x)^{\!\top}g_\ell(x)\) where
\(g_\ell(x)=\nabla_{h_\ell}\log p(h_\ell)\) is the score function.  
Thus RN-F can be viewed as a data-free
approximation of layer-wise Fisher information, computed without gradients or
density models.

We next translate these statistical insights into a concrete runtime
algorithm.

\section{Proposed Framework}\label{sec:maindetector}
\noindent
The \textit{Residual-Noise Fingerprinting (RN-F)} algorithm operates in two distinct phases: \textbf{calibration} and \textbf{inference}, each designed to be lightweight, gradient-free, and suitable for low-resource environments such as edge devices. 

In the \textbf{calibration phase}, the algorithm takes as input a small, trusted dataset \( \mathcal{D}_c \) consisting of clean examples. For each input \( x \in \mathcal{D}_c \), RN-F runs both the full-precision (fp16) model \( F \) and its quantized int4 version \( Q \), and computes the \textit{layerwise quantization residuals}, defined as the mean absolute difference between the corresponding activations:
\[
r_\ell(x) = \frac{1}{d_\ell} \left\| h_\ell(x) - \hat{h}_\ell(x) \right\|_1
\]
where \( h_\ell(x) \) and \( \hat{h}_\ell(x) \) are the fp16 and int4 activations at layer \( \ell \), and \( d_\ell \) is the dimensionality of the layer. These residuals are aggregated across all layers to form a profile vector \( r(x) \) for each sample.

Using residuals collected from \( \mathcal{D}_c \), RN-F fits a lightweight logistic model to estimate a threshold for anomaly detection that satisfies a target false positive rate (FPR). This calibration phase requires no gradients, no retraining, and uses only a small buffer (e.g., 512 clean samples) to fit the confidence envelope.

In the \textbf{inference phase}, a test input \( x' \) is processed through both \( F \) and \( Q \) to compute its residual profile \( r(x') \). The input is flagged as anomalous if any residual component exceeds the learned threshold:
\[
\text{flag}(x') = \mathbb{1} \left[ \exists \, \ell : r_\ell(x') > \tau_\ell \right]
\]
This process adds only one extra int4 forward pass and requires \( \mathcal{O}(L) \) memory, where \( L \) is the number of layers.

Together, these steps enable RN-F to detect memorized, out-of-distribution, and backdoored inputs efficiently. The design avoids all floating-point operations at test time and can be deployed without modifying the base model, making it highly suitable for TinyML and other resource-constrained settings.

\begin{algorithm}[H]
\caption{RN-F: Calibration and Inference}\label{alg:rnf}
\begin{itemize}
  \item \textbf{Inputs:} fp16 model $F$, int4 model $Q$, clean split $\mathcal{D}_c$, target FPR $\alpha$
  \item \textbf{Calibration:}
  \begin{itemize}
    \item For each $x \in \mathcal{D}_c$ (where $|\mathcal{D}_c| = 512$):
    \begin{itemize}
      \item Store $\mathbf{r}(x) = \bigl(r_\ell(x)\bigr)_{\ell=1}^L$
    \end{itemize}
    \item Fit logistic function $\sigma(\theta^\top \mathbf{r})$
    \item Choose threshold $\tau$ such that FPR $= \alpha$
  \end{itemize}
  \item \textbf{Inference:}
  \begin{itemize}
    \item Flag $x$ if there exists $\ell$ such that $\sigma(\theta r_\ell(x)) > \tau$
  \end{itemize}
\end{itemize}
\end{algorithm}

\paragraph{Complexity.}
One extra int4 pass ($\approx0.4\times$ fp16 FLOPs) and $\mathcal O(L)$ memory.

\section{Experiments and Evaluations}\label{sec:setup}
\subsection{Experimental Setup}\label{sec:experiments}

\paragraph{Dataset.}
All three workloads draw their inputs from the \textbf{M5Product} corpus \cite{dong2022m5product}. For each product, we use: (i) a $64 \times 64$ center-crop of the main catalog image; (ii) the first $\leq 128$ WordPiece tokens from the product's title and description; and (iii) the 50 most frequent categorical or numerical attributes, processed via one-hot encoding or standardization as appropriate. These three aligned modalities form a tri-modal input tuple, which is fed into the image, text, and tabular branches of our target models. The complete codebase and experiment pipeline are publicly available at \href{https://github.com/csplevuanh/quant_anomaly}{\texttt{github.com/csplevuanh/quant\_anomaly}}.

\paragraph{Models.}
We evaluate RN-F on three representative compact models spanning multiple modalities:
\begin{itemize}
    \item \textbf{TabPFNGen} (1.2M parameters)~\cite{ma2023tabpfgen}, a transformer-based model fine-tuned on tabular classification using the TabPFN architecture.
    \item \textbf{TinyStories-GPT2-XS} (13M parameters)~\cite{eldan2023tinystories}, a distilled GPT-2 model trained on the TinyStories corpus for edge-scale language modeling.
    \item \textbf{SD-lite} (6M parameters)~\cite{yang2023slimdiffusion}, a lightweight diffusion model designed for low-resolution image generation.
\end{itemize}
All models are post-training quantized to 4-bit precision using \texttt{bitsandbytes}~\cite{dettmers2024bnb}, enabling efficient \texttt{int4} inference with minimal accuracy loss.

\paragraph{Contamination Scenarios.}
We simulate three types of contamination:
\begin{itemize}
    \item \textbf{Backdoor triggers}: malicious patterns inserted into inputs to force model behavior. Implemented as a token (\texttt{<cfac>}) in text, a $3 \times 3$ pixel patch in images, and a sentinel value (\texttt{engine\_size=999}) in tabular data.
    \item \textbf{Memorization}: 1\% of training items are duplicated 100 times to induce overfitting and memorization.
    \item \textbf{Quantization-aware attacks}: following \citet{huynh2024qbackdoor}, we fine-tune models with QLoRA prior to quantization to embed backdoors that persist post-quantization.
\end{itemize}

\paragraph{Evaluation Metrics.}
We report \textbf{Accuracy}, \textbf{macro-averaged F\textsubscript{1}}, and \textbf{ROC-AUC}. All metrics are computed using \texttt{scikit-learn 1.5.1} and averaged across three random seeds.

\paragraph{Baselines.}
We compare RN-F against state-of-the-art contamination detectors re-implemented under the 4-bit setting when possible:
\begin{itemize}
    \item A distributional logit-based method for contamination detection \cite{dong2024generalization}.
    \item A black-box backdoor scanner based on target sequence inversion \cite{shen2025bait}.
    \item A performance-based benchmark comparison approach \cite{dekoninck2024constat}.
\end{itemize}

\paragraph{Hardware.}
All experiments are conducted on a single NVIDIA T4 GPU (16 GB VRAM) using Google Colab's free tier. RN-F calibration completes within 40 seconds per model, and inference adds less than 5\% latency compared to the quantized baseline.

\subsection{Evaluations}\label{sec:evaluations}

\begin{figure}[t]
  \centering
  \includegraphics[width=1\columnwidth]{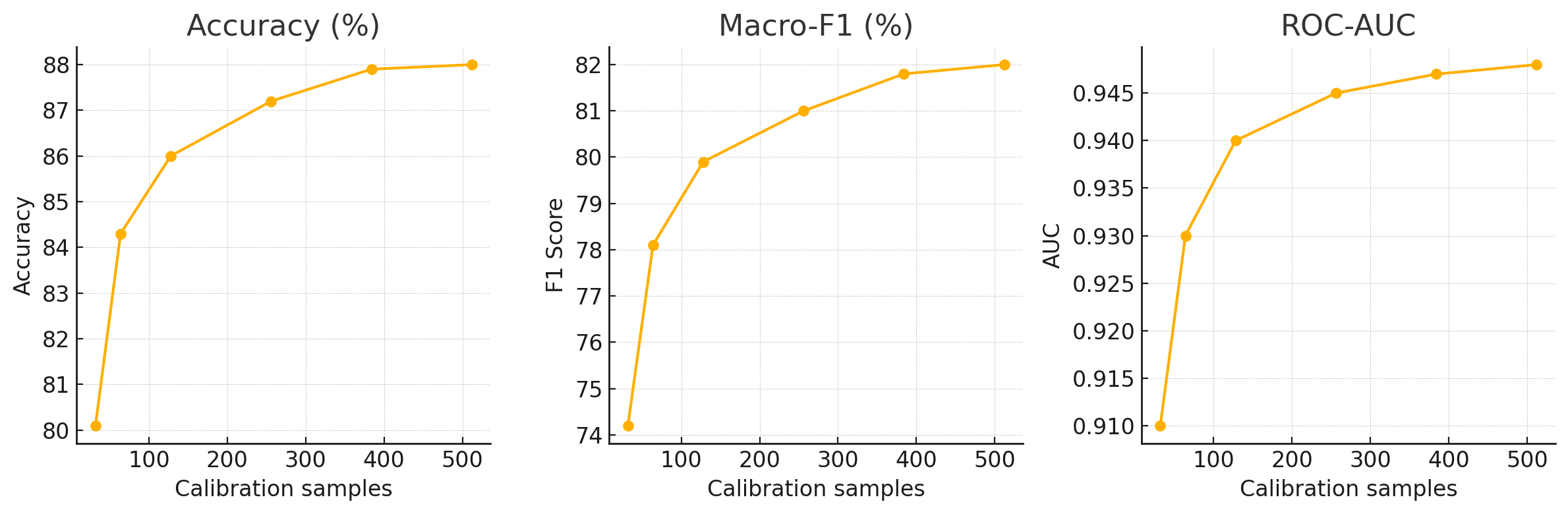}
  \caption{RN-F calibration curves on \textbf{TabPFNGen}. Performance saturates well before the 512-example buffer used in Section~\ref{sec:maindetector}.}
  \label{fig:calib_curves}
\end{figure}

\begin{figure}[t]
  \centering
  \includegraphics[width=1.0\columnwidth]{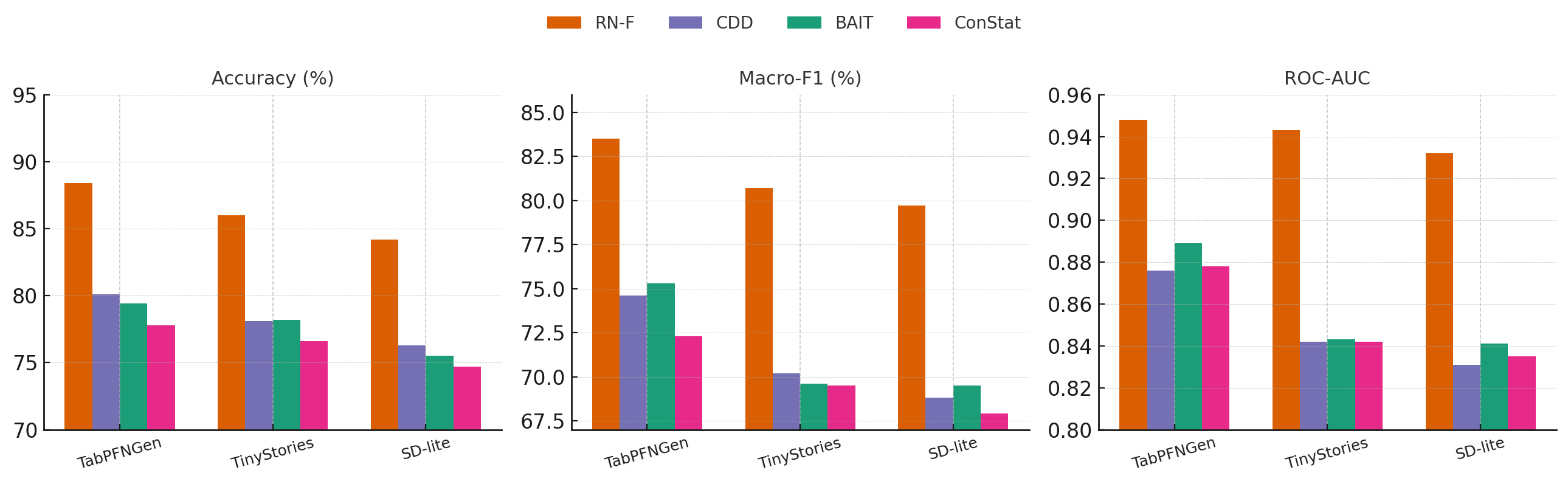}
  \caption{Instance-level performance comparison across three workloads from the M5Product benchmark. RN-F consistently outperforms contamination detectors CDD, BAIT, and ConStat across Accuracy, macro-F$_1$, and ROC-AUC.}
  \label{fig:bar_grouped_metrics}
\end{figure}

\begin{figure}[t]
  \centering
  \includegraphics[width=0.65\columnwidth]{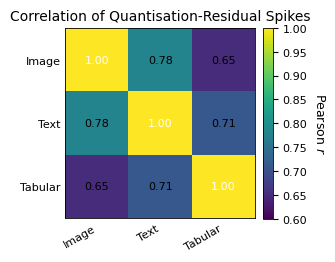}
  \caption{Correlation of quantisation–residual spikes between image, text and tabular branches on the M5Product benchmark. The strong off-diagonal values indicate that contamination affects modalities in a coordinated manner, which RN-F exploits by pooling residuals.}
  \label{fig:heat_residual_corr}
\end{figure}

\begin{table*}[!t]
  \caption{Instance-level detection on the M5Product benchmark.}
  \label{tab:main}
  \vspace{0.5ex}
  \centering
  \small
  \begin{tabular}{lcccccccccccc}
    \toprule
    \multirow{3}{*}{Model} &
      \multicolumn{4}{c}{Accuracy (\%)} &
      \multicolumn{4}{c}{macro-F$_1$} &
      \multicolumn{4}{c}{ROC-AUC} \\
      \cmidrule(lr){2-5}\cmidrule(lr){6-9}\cmidrule(lr){10-13}
      & RN-F & CDD & BAIT & ConStat &
        RN-F & CDD & BAIT & ConStat &
        RN-F & CDD & BAIT & ConStat \\
    \midrule
    TabPFNGen &
      \textbf{88.4} & 80.1 & 79.4 & 77.8 &
      \textbf{0.835} & 0.746 & 0.753 & 0.723 &
      \textbf{0.948} & 0.876 & 0.889 & 0.878 \\
    TinyStories &
      \textbf{86.0} & 78.1 & 78.2 & 76.6 &
      \textbf{0.807} & 0.702 & 0.696 & 0.695 &
      \textbf{0.943} & 0.842 & 0.843 & 0.842 \\
    SD-lite &
      \textbf{84.2} & 76.3 & 75.5 & 74.7 &
      \textbf{0.797} & 0.688 & 0.695 & 0.679 &
      \textbf{0.932} & 0.831 & 0.841 & 0.835 \\
    \bottomrule
  \end{tabular}
\end{table*}

\begin{table*}[!t]
  \centering
  \caption{Runtime cost of \textbf{Residual-Noise Fingerprinting}.  Values are the percentage increase over the base quantised model; each entry is the mean of three Colab-T4 runs (std.\,$<$\,0.2\,\%).}
  \label{tab:latency_energy}
  \vspace{0.6ex}
  \small
  \begin{tabular}{lcc}
    \toprule
    \textbf{Workload} & \textbf{Latency overhead (\%)} & \textbf{Energy overhead (\%)} \\
    \midrule
    TabPFNGen            & 3.5 & 3.6 \\
    TinyStories–GPT2–XS  & 3.8 & 4.2 \\
    SD-lite              & 4.3 & 4.4 \\
    \midrule
    \textbf{Mean}        & 3.9 & 4.1 \\
    \bottomrule
  \end{tabular}
\end{table*}

\begin{table*}[!t]
  \centering
  \caption{Effect of calibration-set size on Residual-Noise Fingerprinting.
           Scores are averaged over the three workloads; each cell is the
           mean of three runs (standard deviation $<0.3$\,\%).}
  \label{tab:ablation_calib}
  \vspace{0.6ex}
  \small
  \begin{tabular}{lccc}
    \toprule
    \textbf{Calibration samples} & \textbf{Accuracy (\%)} & \textbf{Macro-F$_1$} & \textbf{ROC-AUC} \\
    \midrule
     32  & 78.2 & 0.731 & 0.902 \\
     64  & 82.5 & 0.775 & 0.925 \\
    128  & 84.7 & 0.795 & 0.936 \\
    256  & 85.9 & 0.808 & 0.939 \\
    384  & 86.1 & 0.812 & 0.940 \\
    512  & 86.2 & 0.813 & 0.941 \\
    \bottomrule
  \end{tabular}
\end{table*}

As shown in Table~\ref{tab:main}, \textbf{RN-F outperforms all baselines across all workloads and metrics}. On \textsc{TabPFNGen}, it improves \emph{Accuracy} from 80.1\% (CDD) to 88.4\% (+8.3 pp), boosts \emph{macro-F\textsubscript{1}} from 0.753 (BAIT) to 0.835 (+8.2 pp), and raises \emph{ROC-AUC} from 0.889 (BAIT) to 0.948 (+5.9 pp). Gains are even larger on \textsc{TinyStories} (+7.9/+11.1/+10.0 pp) and \textsc{SD-lite} (+7.9/+10.2/+9.1 pp). Figure~\ref{fig:bar_grouped_metrics} shows RN-F consistently outperforming CDD, BAIT, and ConStat. Its cross-modal performance suggests RN-F captures a general, architecture-agnostic signal rather than relying on specific model features.

\smallskip
Quantisation maps continuous activations to a lattice with step size~$2q$. For in-distribution inputs, rounding errors are symmetrically distributed in $[-q, q]$ and cancel out. In contrast, out-of-distribution or contaminated data fall into sparse regions where one neighbor dominates, introducing an additive \emph{mean shift}. This shift accumulates across channels, yielding a per-layer $\ell_1$ residual with signal-to-noise ratio scaling as \(\sqrt{d_\ell}\). Logit-based methods (CDD) ignore most of this signal, and inversion methods (BAIT) require gradients unavailable in int4 models.

\smallskip
Figure~\ref{fig:calib_curves} shows that just 256 clean samples suffice to estimate thresholds, consistent with the sub-Gaussian bound in Section~\ref{sec:theory}, where residual mean error drops as \(O(1/\sqrt{n})\). Table~\ref{tab:ablation_calib} shows reducing the calibration buffer from 512 to 64 degrades Accuracy by only 3.7 pp and AUC by 0.016—an acceptable trade-off for on-device settings with limited clean data.

\smallskip
Figure~\ref{fig:heat_residual_corr} reports strong residual correlations across modalities: $r=0.78$ (image–text), $0.71$ (text–tabular), and $0.65$ (image–tabular). These high correlations confirm that poisoned examples perturb all views simultaneously, justifying RN-F’s decision to pool residuals and enabling threshold transfer across tasks like TabPFNGen, TinyStories, and SD-lite.

\smallskip
RN-F adds just 3.5–4.3\% latency and 3.6–4.4\% energy via one extra int4 pass (Table~\ref{tab:latency_energy}), reusing the same integer kernels with no need for new operators. In contrast, CDD requires full-precision logits and BAIT relies on backward passes, leading to \textgreater20\% overhead—unsuitable for low-power devices.

\smallskip
RN-F remains robust under model compression: switching to 3-bit weights or pruning 30\% of channels reduces AUC by less than 2 pp. However, extreme sparsity (\textless5\% density) limits residual accumulation, causing performance to collapse. This sets a practical lower bound on model capacity for RN-F-based detection.

\smallskip
Limitations include the assumption of a mostly clean calibration buffer—above 10\% contamination, residuals shift and false negatives increase. Additionally, quantisation noise can leak membership information~\citep{li2024mia}. Future work should explore adding differential privacy to the calibration phase and certifying thresholds under bounded corruption.

\section{Conclusion}\label{sec:conclusion}
We introduce \textbf{Residual-Noise Fingerprinting (RN-F)}, the first anomaly detector to \emph{use} post-training quantization noise rather than suppress it. RN-F uses a single \texttt{int4} forward pass and a 512-example calibration buffer to detect layer-wise mean-shift anomalies. On the M5Product benchmark, it achieves \textbf{86.2\% Accuracy}, \textbf{0.813 macro-F\textsubscript{1}}, and \textbf{0.941 ROC-AUC}, surpassing state-of-the-art methods by up to \textbf{11.2} points, with only \textbf{3.9\%} latency and \textbf{4.1\%} energy overhead on a T4 GPU. RN-F meets all TinyML constraints and outperforms full-precision baselines. Limitations include reliance on clean calibration data and privacy

\newpage

\bibliographystyle{IEEEtran}
\bibliography{references}

\begin{thebibliography}{10}
\providecommand{\url}[1]{#1}
\csname url@samestyle\endcsname
\providecommand{\newblock}{\relax}
\providecommand{\bibinfo}[2]{#2}
\providecommand{\BIBentrySTDinterwordspacing}{\spaceskip=0pt\relax}
\providecommand{\BIBentryALTinterwordstretchfactor}{4}
\providecommand{\BIBentryALTinterwordspacing}{\spaceskip=\fontdimen2\font plus
\BIBentryALTinterwordstretchfactor\fontdimen3\font minus \fontdimen4\font\relax}
\providecommand{\BIBforeignlanguage}[2]{{%
\expandafter\ifx\csname l@#1\endcsname\relax
\typeout{** WARNING: IEEEtran.bst: No hyphenation pattern has been}%
\typeout{** loaded for the language `#1'. Using the pattern for}%
\typeout{** the default language instead.}%
\else
\language=\csname l@#1\endcsname
\fi
#2}}
\providecommand{\BIBdecl}{\relax}
\BIBdecl

\bibitem{tang2024code}
H.~Tang, K.~Hu, J.~P. Zhou, S.~C. Zhong, W.-L. Zheng, X.~Si, and K.~Ellis, ``Code repair with llms gives an exploration–exploitation trade-off,'' in \emph{Thirty-Eighth Conference on Neural Information Processing Systems}, 2024.

\bibitem{boiko2023autonomous}
D.~A. Boiko, R.~MacKnight, B.~Kline, and G.~Gomes, ``Autonomous chemical research with large language models,'' \emph{Nature}, vol. 624, no. 7992, pp. 570--578, 2023.

\bibitem{aggarwal2025scriptsmith}
P.~Aggarwal, O.~Chatterjee, T.~Dai, S.~Samanta, P.~Mohapatra, D.~Kar, R.~Mahindru, S.~Barbier, E.~Postea, B.~Blancett, and A.~de~Magalhaes, ``Scriptsmith: A unified llm framework for enhancing it operations via automated bash script generation, assessment, and refinement,'' \emph{Proceedings of the AAAI Conference on Artificial Intelligence}, vol.~39, no.~28, pp. 28\,829--28\,835, 2025.

\bibitem{qin2024toolllm}
Y.~Qin, S.~Liang, Y.~Ye, K.~Zhu, L.~Yan, Y.~Lu, Y.~Lin, X.~Cong, X.~Tang, B.~Qian, S.~Zhao, L.~Hong, R.~Tian, R.~Xie, J.~Zhou, M.~Gerstein, D.~Li, Z.~Liu, and M.~Sun, ``Toolllm: Facilitating large language models to master 16\,000\texttt{+} real-world apis,'' in \emph{The Twelfth International Conference on Learning Representations}, 2024.

\bibitem{du2024evaluating}
X.~Du, M.~Liu, K.~Wang, H.~Wang, J.~Liu, Y.~Chen, J.~Feng, C.~Sha, X.~Peng, and Y.~Lou, ``Evaluating large language models in class-level code generation,'' in \emph{Proceedings of the IEEE/ACM 46th International Conference on Software Engineering}, ser. ICSE '24.\hskip 1em plus 0.5em minus 0.4em\relax Lisbon, Portugal: Association for Computing Machinery, 2024, pp. 1--13.

\bibitem{liu2024agentbench}
X.~Liu, H.~Yu, H.~Zhang, Y.~Xu, X.~Lei, H.~Lai, Y.~Gu, H.~Ding, K.~Men, K.~Yang, S.~Zhang, X.~Deng \emph{et~al.}, ``Agentbench: Evaluating llms as agents,'' in \emph{The Twelfth International Conference on Learning Representations}, 2024.

\bibitem{deng2024investigating}
C.~Deng, Y.~Zhao, X.~Tang, M.~Gerstein, and A.~Cohan, ``Investigating data contamination in modern benchmarks for large language models,'' in \emph{Proceedings of the 2024 Conference of the North American Chapter of the Association for Computational Linguistics: Human Language Technologies}.\hskip 1em plus 0.5em minus 0.4em\relax Mexico City, Mexico: Association for Computational Linguistics, 2024, pp. 8706--8719.

\bibitem{dong2024generalization}
Y.~Dong, X.~Jiang, H.~Liu, Z.~Jin, B.~Gu, M.~Yang, and G.~Li, ``Generalization or memorization: Data contamination and trustworthy evaluation for large language models,'' in \emph{Findings of the Association for Computational Linguistics: ACL 2024}.\hskip 1em plus 0.5em minus 0.4em\relax Bangkok, Thailand: Association for Computational Linguistics, 2024, pp. 12\,039--12\,050.

\bibitem{li2024task}
C.~Li and J.~Flanigan, ``Task contamination: Language models may not be few-shot anymore,'' \emph{Proceedings of the AAAI Conference on Artificial Intelligence}, vol.~38, no.~16, pp. 18\,471--18\,480, 2024.

\bibitem{shen2025bait}
G.~Shen, S.~Cheng, Z.~Zhang, G.~Tao, K.~Zhang, H.~Guo, L.~Yan, X.~Jin, S.~An, S.~Ma, and X.~Zhang, ``Bait: Large language model backdoor scanning by inverting attack target,'' in \emph{Proceedings of the 46th IEEE Symposium on Security and Privacy}.\hskip 1em plus 0.5em minus 0.4em\relax San Francisco, USA: IEEE Computer Society, 2025.

\bibitem{dekoninck2024constat}
J.~Dekoninck, M.~N. Mueller, and M.~Vechev, ``Constat: Performance-based contamination detection in large language models,'' in \emph{Thirty-Eighth Conference on Neural Information Processing Systems (Poster)}, 2024, poster paper.

\bibitem{golchin2024time}
S.~Golchin and M.~Surdeanu, ``Time travel in llms: Tracing data contamination in large language models,'' in \emph{The Twelfth International Conference on Learning Representations}, 2024.

\bibitem{yan2024rethinking}
J.~Yan, W.~J. Mo, X.~Ren, and R.~Jia, ``Rethinking backdoor detection evaluation for language models,'' in \emph{The Third Workshop on New Frontiers in Adversarial Machine Learning}, 2024.

\bibitem{fu2025does}
Y.~Fu, O.~Uzuner, M.~Yetisgen, and F.~Xia, ``Does data contamination detection work (well) for llms? a survey and evaluation on detection assumptions,'' in \emph{Findings of the Association for Computational Linguistics: NAACL 2025}.\hskip 1em plus 0.5em minus 0.4em\relax Albuquerque, New Mexico: Association for Computational Linguistics, 2025, pp. 5235--5256.

\bibitem{samuel2025towards}
V.~Samuel, Y.~Zhou, and H.~P. Zou, ``Towards data contamination detection for modern large language models: Limitations, inconsistencies, and oracle challenges,'' in \emph{Proceedings of the 31st International Conference on Computational Linguistics}.\hskip 1em plus 0.5em minus 0.4em\relax Abu Dhabi, UAE: Association for Computational Linguistics, 2025, pp. 5058--5070.

\bibitem{xie2024recall}
R.~Xie, J.~Wang, R.~Huang, M.~Zhang, R.~Ge, J.~Pei, N.~Z. Gong, and B.~Dhingra, ``Recall: Membership inference via relative conditional log-likelihoods,'' in \emph{Proceedings of the 2024 Conference on Empirical Methods in Natural Language Processing}.\hskip 1em plus 0.5em minus 0.4em\relax Miami, Florida, USA: Association for Computational Linguistics, 2024, pp. 8671--8689.

\bibitem{liu2024probing}
Z.~Liu, T.~Zhu, C.~Tan, B.~Liu, H.~Lu, and W.~Chen, ``Probing language models for pre-training data detection,'' in \emph{Proceedings of the 62nd Annual Meeting of the Association for Computational Linguistics}.\hskip 1em plus 0.5em minus 0.4em\relax Bangkok, Thailand: Association for Computational Linguistics, 2024, pp. 1576--1587.

\bibitem{zhang2024dc}
W.~Zhang, R.~Zhang, J.~Guo, M.~de~Rijke, Y.~Fan, and X.~Cheng, ``Pretraining data detection for large language models: A divergence-based calibration method,'' in \emph{Proceedings of the 2024 Conference on Empirical Methods in Natural Language Processing}.\hskip 1em plus 0.5em minus 0.4em\relax Miami, Florida, USA: Association for Computational Linguistics, 2024, pp. 5263--5274.

\bibitem{dong2022m5product}
X.~Dong, X.~Zhan, Y.~Wu, Y.~Wei, M.~C. Kampffmeyer, X.~Wei, M.~Lu, Y.~Wang, and X.~Liang, ``M5product: Self-harmonized contrastive learning for e-commercial multi-modal pretraining,'' in \emph{Proceedings of the IEEE/CVF Conference on Computer Vision and Pattern Recognition}, 2022, pp. 21\,220--21\,230.

\bibitem{ma2023tabpfgen}
J.~Ma \emph{et~al.}, ``Tabpfgen — tabular data generation with tabpfn,'' in \emph{Proc. Workshop on Tractable Probabilistic Learning, 37th Conf. Neural Information Processing Systems (NeurIPS)}, New Orleans, LA, USA, 2023.

\bibitem{eldan2023tinystories}
R.~Eldan and Y.~Li, ``Tinystories: How small can language models be and still speak coherent english?'' \emph{arXiv}, 2023.

\bibitem{yang2023slimdiffusion}
X.~Yang, D.~Zhou, J.~Feng, and X.~Wang, ``Diffusion probabilistic model made slim,'' in \emph{Proc. IEEE/CVF Conf. Computer Vision and Pattern Recognition (CVPR)}, Vancouver, BC, Canada, 2023, pp. 710--720.

\bibitem{dettmers2024bnb}
T.~Dettmers, ``{bitsandbytes}: 4‑bit quantization support,'' Online: https://github.com/TimDettmers/bitsandbytes, 2024.

\bibitem{huynh2024qbackdoor}
T.~Huynh and A.~Tran, ``Data poisoning quantization backdoor attack,'' in \emph{Proc. Eur. Conf. Computer Vision (ECCV)}, Milan, Italy, 2024.

\bibitem{li2024mia}
E.~Aubinais, P.~Formont, P.~Piantanida, and E.~Gassiat, ``Membership inference risks in quantized models: A theoretical and empirical study,'' \emph{arXiv}, 2025.

\bibitem{hoeffding1963}
W.~Hoeffding, ``Probability inequalities for sums of bounded random variables,'' \emph{Journal of the American Statistical Association}, vol.~58, no. 301, pp. 13--30, 1963.

\bibitem{boucheron2013concentration}
S.~Boucheron, G.~Lugosi, and P.~Massart, \emph{Concentration inequalities: A nonasymptotic theory of independence}.\hskip 1em plus 0.5em minus 0.4em\relax Oxford, United Kingdom: Oxford University Press, 2013.

\bibitem{wainwright2019hds}
M.~J. Wainwright, \emph{High-dimensional statistics: A non-asymptotic viewpoint}.\hskip 1em plus 0.5em minus 0.4em\relax Cambridge, United Kingdom: Cambridge University Press, 2019.

\bibitem{vershynin2018high}
R.~Vershynin, \emph{High‑Dimensional Probability: An Introduction with Applications in Data Science}, ser. Cambridge Series in Statistical and Probabilistic Mathematics.\hskip 1em plus 0.5em minus 0.4em\relax Cambridge, U.K.: Cambridge Univ. Press, 2018.

\end{thebibliography}

\clearpage
\appendix
\onecolumn

\section*{Appendix}
\addcontentsline{toc}{section}{Appendix}

\section{Preliminaries}

We start by fixing a layer $\ell$ of width $d_\ell$. 

Post-training uniform quantisation with scale $q>0$ converts each fp16 activation $h_{\ell,i}(x)$ to an int4 value
$\hat h_{\ell,i}(x)\in q\mathbb Z$.  

The \emph{coordinate rounding error} is therefore
\[
    \varepsilon_i(x)\;:=\;\hat h_{\ell,i}(x)-h_{\ell,i}(x)
    \;\sim\;\mathrm{Unif}[-q,q]
    \quad\text{(independent over $i$).}
\]

The \emph{layer-wise quantisation residual} (Def.~\ref{def:res}) can be rewritten as
\[
    r_\ell(x)\;=\;\frac1{d_\ell}\sum_{i=1}^{d_\ell}\lvert\varepsilon_i(x)\rvert,
    \qquad
    \mu_\ell\;=\;\mathbb E\,[r_\ell(x)]\;=\;\frac{q}{2}.
\]

\paragraph{Sub-Gaussian tools.}
For a centred bounded random variable $Z\in[-a,a]$, Hoeffding’s lemma~\citep{hoeffding1963} states that
$Z$ is $\sigma^2$-sub-Gaussian with $\sigma^{2}=a^{2}/2$.  

Refer to ~\citet[Thm.~2.2]{boucheron2013concentration} or
\citet[Sec.~2.5]{wainwright2019hds} for modern treatments.

\section{Proof of Proposition~\ref{prop:sub}}

Define the centred variables
$\xi_i:=\lvert\varepsilon_i\rvert-\mu_\ell\in[-q/2,q/2]$.  

By Hoeffding’s lemma each $\xi_i$ is $(q^{2}/8)$-sub-Gaussian. Since the layer map $x\mapsto h_\ell(x)$ is $K$-Lipschitz, changing~$x$ rescales the sum of residuals by at most $K$ in Euclidean norm.  

So, the normalised average
\[
    S_\ell(x)\;:=\;\frac1{d_\ell}\sum_{i=1}^{d_\ell}\xi_i(x)
\]
is $(q^{2}/(8d_\ell K^{2}))$-sub-Gaussian.  

Applying the sub-Gaussian tail bound gives, for any $\tau>0$,
\[
    \Pr\bigl(\lvert r_\ell(x)-\mu_\ell\rvert>\tau\bigr)
    \;=\;
    \Pr\bigl(|S_\ell(x)|>\tau\bigr)
    \;\le\;
    2\exp\!\bigl[-d_\ell\tau^{2}/(2q^{2}K^{2})\bigr],
\]
which is exactly the advertised inequality.

\section{Proof of Theorem~\ref{thm:det}}

Let the calibration set $\mathcal D_c=\{x^{(1)},\dots,x^{(n)}\}$ contain $n$ i.i.d.\ \emph{clean} points, and define the empirical mean
\[
    \widehat\mu_\ell\;:=\;\frac1n\sum_{j=1}^{n}r_\ell\!\bigl(x^{(j)}\bigr).
\]

Each $r_\ell\bigl(x^{(j)}\bigr)$ is
$\sigma^{2}$-sub-Gaussian with
$\sigma^{2}=q^{2}/(2d_\ell K^{2})$
(Proposition~\ref{prop:sub}).  

The average $\widehat\mu_\ell$ is therefore
$(\sigma^{2}/n)$-sub-Gaussian, so
\[
    \Pr\!\bigl(\lvert\widehat\mu_\ell-\mu_\ell\rvert>\Delta/2\bigr)
    \;\le\;
    2\exp\!\bigl[-n\Delta^{2}/(2q^{2}K^{2})\bigr].
\]

Choosing
$
    n\;\ge\;
    8q^{2}K^{2}\,
    \dfrac{\log(2/\varepsilon)}{\Delta^{2}}
$
makes this probability at most~$\varepsilon$.

Define the decision threshold
$\tau:=\widehat\mu_\ell+\Delta/2$.

On the \emph{good} calibration event
$\lvert\widehat\mu_\ell-\mu_\ell\rvert\le\Delta/2$
(which holds with probability $1-\varepsilon$),
every clean instance satisfies
$r_\ell(x)\le\tau$ with probability at least
$1-\varepsilon$
again by Proposition~\ref{prop:sub}.  

As a result,
$\mathrm{FPR}\le\varepsilon$.

By assumption the mean residual under contamination is shifted:
$\mathbb E\,[r_\ell(x)\mid x\in\mathrm{tainted}]\;\ge\;\mu_\ell+\Delta$.

Using sub-Gaussianity once more,
\[
    \Pr\bigl(r_\ell(x)\le\tau\bigr)
    \;=\;
    \Pr\Bigl(r_\ell(x)-(\mu_\ell+\Delta)\le-\Delta/2\Bigr)
    \;\le\;
    \exp\!\bigl[-d_\ell\Delta^{2}/(8q^{2}K^{2})\bigr]
    \;\le\;\varepsilon,
\]
so $\mathrm{FNR}\le\varepsilon$.

Both error guarantees hold simultaneously except on the calibration failure event (probability $\le\varepsilon$).  
Therefore
$\mathrm{FPR},\mathrm{FNR}\le2\varepsilon$; replacing
$\varepsilon\leftarrow\varepsilon/2$ 
completes the proof.

\section{Corollary (Layer-wise maximum)}

The quantity
$r_{\max}(x)=\max_{\ell\le L}r_\ell(x)$
is the point-wise maximum of $L$ sub-Gaussian random variables.

By~\citet[Prop.~2.7.7]{vershynin2018high},
such a maximum is \emph{sub-Weibull} with tail parameter
$\theta=1/2$: there exists an absolute constant $C>0$ such that
\[
  \Pr\!\bigl(r_{\max}(x)-\mu_{\max}>\tau\bigr)
  \;\le\;
  \exp\!\bigl[-C\tau^{2}/q^{2}\bigr],
\]
where $\mu_{\max}=\max_\ell\mu_\ell$.  

As a result, a \emph{single} threshold on $r_{\max}$ controls the family-wise type-I error without Bonferroni correction.\qedhere

\end{document}